\documentclass[runningheads]{llncs}

\usepackage[T1]{fontenc}
\usepackage[hyphens]{url}
\usepackage{graphicx}
\usepackage[T1]{fontenc}
\usepackage{booktabs}
\begin{document}
\title{Physiological and Semantic Patterns in Medical Teams Using an Intelligent Tutoring System}
\titlerunning{Physiological and Semantic Patterns in an Intelligent Tutoring System}
\author{Xiaoshan Huang\inst{1}\orcidID{0000-0002-2853-7219} \and
Conrad Borchers\inst{2}\orcidID{0000-0003-3437-8979} \and
Jiayi Zhang\inst{3}\orcidID{0000-0002-7334-4256} \and Susanne P. Lajoie\inst{1}\orcidID{0000-0003-2814-3962}}
\authorrunning{X. Huang et al.}

\institute{McGill University, Montreal, Canada\\
\email{xiaoshan.huang@mail.mcgill.ca, susanne.lajoie@mcgill.ca} \and
Carnegie Mellon University, Pittsburgh, USA\\
\email{cborcher@cs.cmu.edu} \and
University of Pennsylvania, Philadelphia, USA\\
\email{joycez@upenn.edu}}

\maketitle              %
\begin{abstract}
Effective collaboration requires teams to manage complex cognitive and emotional states through Socially Shared Regulation of Learning (SSRL). Physiological synchrony (i.e., longitudinal alignment in physiological signals) can indicate these states, but is hard to interpret on its own. We investigate the physiological and conversational dynamics of four medical dyads diagnosing a virtual patient case using an intelligent tutoring system. Semantic shifts in dialogue were correlated with transient physiological synchrony peaks. We also coded utterance segments for SSRL and derived cosine similarity using sentence embeddings. The results showed that activating prior knowledge featured significantly lower semantic similarity than simpler task execution. High physiological synchrony was associated with lower semantic similarity, suggesting that such moments involve exploratory and varied language use. Qualitative analysis triangulated these synchrony peaks as ``pivotal moments'': successful teams synchronized during shared discovery, while unsuccessful teams peaked during shared uncertainty. This research advances human-centered AI by demonstrating how biological signals can be fused with dialogues to understand critical moments in problem solving.

\end{abstract}

\section{Introduction \& Related Work}

Collaborative problem solving in intelligent tutoring systems (ITS) requires learners to jointly manage task-related cognition, motivation, emotion, and interaction~\cite{huang2025makes}. Socially Shared Regulation of Learning (SSRL) provides a theoretical lens for understanding how groups collectively monitor, evaluate, and adapt their learning during such interactions \cite{jarvela2013new}. However, detecting when SSRL occurs remains a central challenge for AI in education (AIED), particularly in autonomous collaborative settings. 
Recent advances in multimodal learning analytics (MMLA) have enabled intelligent systems to capture learners' physiological signals alongside interaction data, offering new opportunities to infer latent regulatory processes \cite{howley2016towards}. Rather than sustained averages of heart rates, transient peaks and change points appear to mark moments of team engagement\cite{huang2025examining}. Crucially, these physiological markers provide an objective window of emotions in teamwork.
At the same time, progress in large language model (LLM) has modeled semantic dynamics in dialogue using sentence embeddings. Prior work has demonstrated that semantic convergence and divergence in dialogue are associated with shared understanding, conceptual change, and coordination among collaborators~\cite{martinez2021you}. Yet, despite their promise, semantic models have rarely been integrated with physiological measures to examine how meaning-making and embodied coordination unfold together during collaboration.

\subsection{Semantic Embeddings for Cognitive Processes}
Embeddings are a critical component of LLM, that provide semantic representations of texts as high-dimensional vectors \cite{koroteev2021bert}. These embeddings encode both the meanings of individual words and the relationships between them, enabling more nuanced and context-sensitive representations of text. Specifically, sentence embeddings have been increasingly used to analyze student-generated text for a variety of purposes, ranging from analyzing and scoring written responses \cite{borchers2025disentangling}, to understanding tutoring dialogue \cite{carmon2023assessment}, and measuring learning and engagement \cite{chen2024multi} in both individual and collaborative learning environments~\cite{zhang2026using}. Despite preliminary success, the ability to detect cognitive processes depends on the constructs (which may vary in concreteness and level of abstraction; \cite{liu2024assessing}) as well as the availability of additional sources of data to contextualize the utterances \cite{zhang2026using}.

\subsection{Physiological Synchrony for Shared Affective Engagement}
Physiological synchrony (PS), the temporal alignment of biological signals between individuals, has emerged as a salient marker for shared affective and cognitive states in collaborative learning~\cite{malmberg2019we}. When learners are deeply engaged in a joint task, their autonomic nervous systems often synchronize, reflecting a shared emotional resonance and mutual regulatory effort~\cite{liu2021collaborative}. 

In SSRL, cognitive interaction involves monitoring task progress, reflection, and the active processing of information. Metacognitive interaction involves the group's collective ``thinking about thinking''. Motivational and affective interactions are deeply intertwined with the group's cognitive dimension. In this study, we focus on physiological synchrony peaks as candidate moments of shared regulatory activity and analyze how these peaks align with semantic shifts in dialogue, SSRL behaviors, and learners’ subjective collaboration experiences. Specifically, we address: \textbf{RQ1}. How semantic shifts in dialogue relate to physiological synchrony and SSRL interactions; and \textbf{RQ2}. how teams experience pivotal collaborative moments.

\section{Methodology}
\subsection{Study Context and Sample}
In this study, we used BioWorld, an ITS designed to foster medical practitioners' clinical reasoning~\cite{huang2023exploring,huang2025linking}. The interface facilitates an iterative problem-solving process through four core functions: (1)Patient Case: Learners highlight critical symptoms from the patient profile. (2) Tests: Teams search for and order diagnostic tests, receiving results instantly. (3) Library: A resource for verifying medical information and supporting knowledge acquisition. (4) Hypotheses: Learners evaluate and rank evidence for various diagnoses. Upon final submission, the system provides an expert report for teams to reflect on their performance.

The IRB-approved study involved medical residents pursuing a Doctor of Medicine at a large North American university. The participants include four dyads of female medical professionals with an average age 28.58 years (\textit{SD} = 2.84) who signed the consent form. Each dyad shared a single computer to discuss and solve the case collaboratively. During the task, the participants' dialogue was collected through audio recordings and their physiological signals through Empatica E4 bracelets, a watch-like device worn on their non-dominant hand. After task completion, we conducted semi-structured interviews with each participant about their perceived challenges and emotional feelings during collaboration.

\subsection{Data Processing and Analyses}
To prepare the multimodal dataset for analysis, verbal and physiological data were processed. Audio recordings of team dialogues and interviews were transcribed using OpenAI Whisper. To ensure transcription accuracy and coding reliability, two researchers validated the text and solved any discrepancies. For the integration of physiological states, the average heart rate was derived from raw blood volume pulse (BVP) signals. These signals were processed at a default sample rate of 1 Hz to maintain temporal alignment with team actions. 

\subsubsection{Semantic Analysis and PS.}
We examined how moments of PS are associated with dialogue semantics. Each dialogue was segmented at the utterance level, and was represented using sentence-level embeddings. These embeddings are numerical representations of entire utterances produced by a pretrained language model, reflecting semantic meaning, including relationships between words and their usage in context. As a result, sentences with similar meanings are mapped to nearby locations in the embedding space. Separately, PS was measured continuously over time and then aligned to each transcript segment.

For every transcript segment, we computed several summaries of PS over the time window corresponding to that segment. These summaries included:
(1) the average synchrony across the entire window,
(2) the minimum synchrony observed in the window, and
(3) the maximum synchrony, or the single highest synchrony value observed within the window. This maximum value, or ``PS peak,'' captures brief, transient spikes in PS rather than sustained alignment.

To test whether language differs during moments of PS peak, we focused on extreme values. For a range of thresholds (the top 1\%, 5\%, 10\%, up to 50\% of synchrony values), we identified transcript segments for which the synchrony sum exceeded the corresponding threshold. For each threshold, we computed the average language embedding of these high-synchrony segments and compared it to the average embedding of all remaining segments. The difference between the two groups was quantified using cosine similarity between embedding centroids. Statistical significance was evaluated using a permutation-based randomization test to evaluate semantic separation (see \cite{borchers2025can}), using 5,000 permutations and a Benjamini–Hochberg correction ($\alpha=0.05$).

\subsubsection{Mixed-Method Bio-Semantic Mapping.} We coded Socially Shared Regulation of Learning (SSRL) behaviors into micro regulatory interactions of meta-cognitive, cognitive, emotional, and motivational processes~\cite{huang2025examining}. These behaviors were analyzed in relation to intra-group physiological synchrony (PS) and semantic variation in dialogue, examining both within- and between-speaker patterns. High-synchrony segments were then qualitatively reviewed to identify underlying patterns. Themes were developed by triangulating task dialogue with semi-structured interviews, incorporating participants’ reflections to validate coordination peaks from physiological data. Interview analysis further revealed themes of shared exploration, emotional states, and perceived collaborative challenges. %

\section{Results}

\subsection{Mapping Physiological and Semantic Patterns with SSRL}
The results showed a clear and interpretable pattern. When PS was summarized using average or minimum values, no reliable semantic differences emerged between high-coordination segments and the rest of the dialogue. In contrast, strong effects appeared when synchrony was indexed by maximum values. %
Transcript segments containing such PS peaks showed consistently greater semantic separation from the rest of the dialogue. These effects were observed across multiple thresholds (for example, when defining high coordination as the top 10–30\% of peak values) and remained statistically significant after correction. Across all tests, 11 out of 66 survived correction, and all significant effects involved the maximum-based coordination measure.

We then investigated the relationship between conversational similarity, bio-measures, and SSRL. We examined two distinct types of linguistic similarity: within-speaker similarity (internal consistency of individuals' language use) and between-speaker similarity (consistency between speakers within the same dyads). %
The result revealed several key patterns. First, both within-speaker and between-speaker similarity showed significant variation across different interactions. For within-speaker similarity, an ANOVA showed significant differences across SSRL codes ($F(4, 249) = 11.538, p < .001$). Similarly, between-speaker similarity also varied significantly by SSRL code ($F(4, 249) = 11.056, p < .001$).

The mixed-effects models revealed a surprising pattern regarding synchrony. Higher PS was associated with significantly lower within-speaker similarity ($\beta = -0.011, p = .001$) and lower between-speaker similarity ($\beta = -0.011, p = .001$). This suggests that high PS moments correspond with less predictable language use both within individual speakers and between conversation partners. Specific interactive behaviors showed distinctive patterns. Task Execution (TE) was consistently associated with higher language similarity, showing significantly greater within-speaker consistency ($\beta = 0.029, p = .011$) and between-speaker alignment ($\beta = 0.039, p = .001$). This pattern suggests that team's task executive actions feature more predictable and aligned language patterns. Social emotional support (SES) also showed higher similarity for both within-speaker ($\beta = 0.044, p = .001$) and between-speaker ($\beta = 0.028, p = .036$) measures. In contrast, Activating prior knowledge (PKA) showed lower within-speaker similarity ($\beta = -0.034, p = .005$) and marginally lower between-speaker similarity ($\beta = -0.023, p = .050$), indicating more varied language use.

Pairwise comparisons revealed that TE showed significantly higher between-speaker similarity than PKA ($t = 6.33, p < .001$), TR ($t = 3.93, p < .001$), and CU ($t = 3.92, p < .001$). For within-speaker similarity, TE showed significantly higher consistency than TPL ($t = 4.64, p < .001$), and PKA showed significantly lower consistency than TE ($t = -4.07, p < .001$). %

\begin{figure}[h]
\centering
\includegraphics[width=\linewidth]{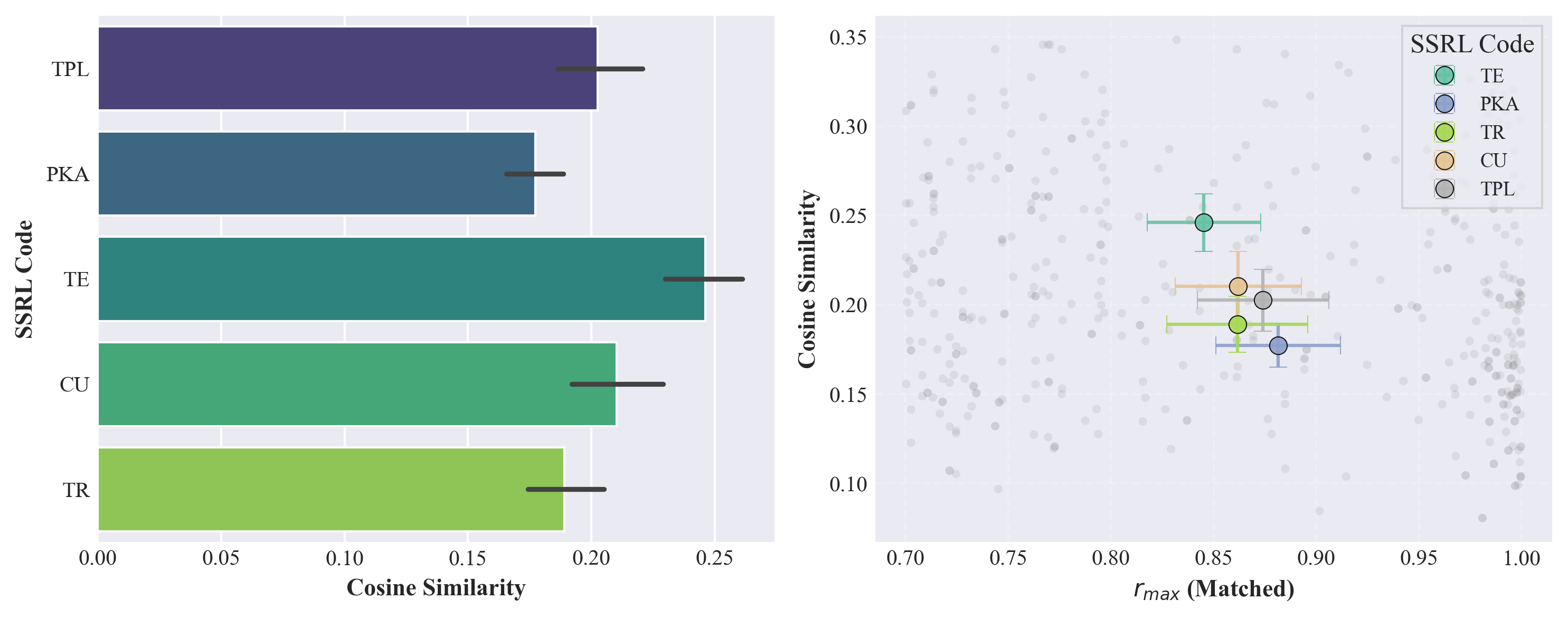}
\caption{Within-speaker similarity patterns. Left: Mean cosine similarity (±95\% CI) for the five most frequent SSRL codes, revealing systematic differences in internal consistency across interaction types. Right: Association between physiological synchrony peaks ($r_{max}$) and cosine similarity, with SSRL code means and 95\% CI error bars.}
\label{fig:within_similarity}
\end{figure}

\begin{figure}[h]
\centering
\includegraphics[width=\linewidth]{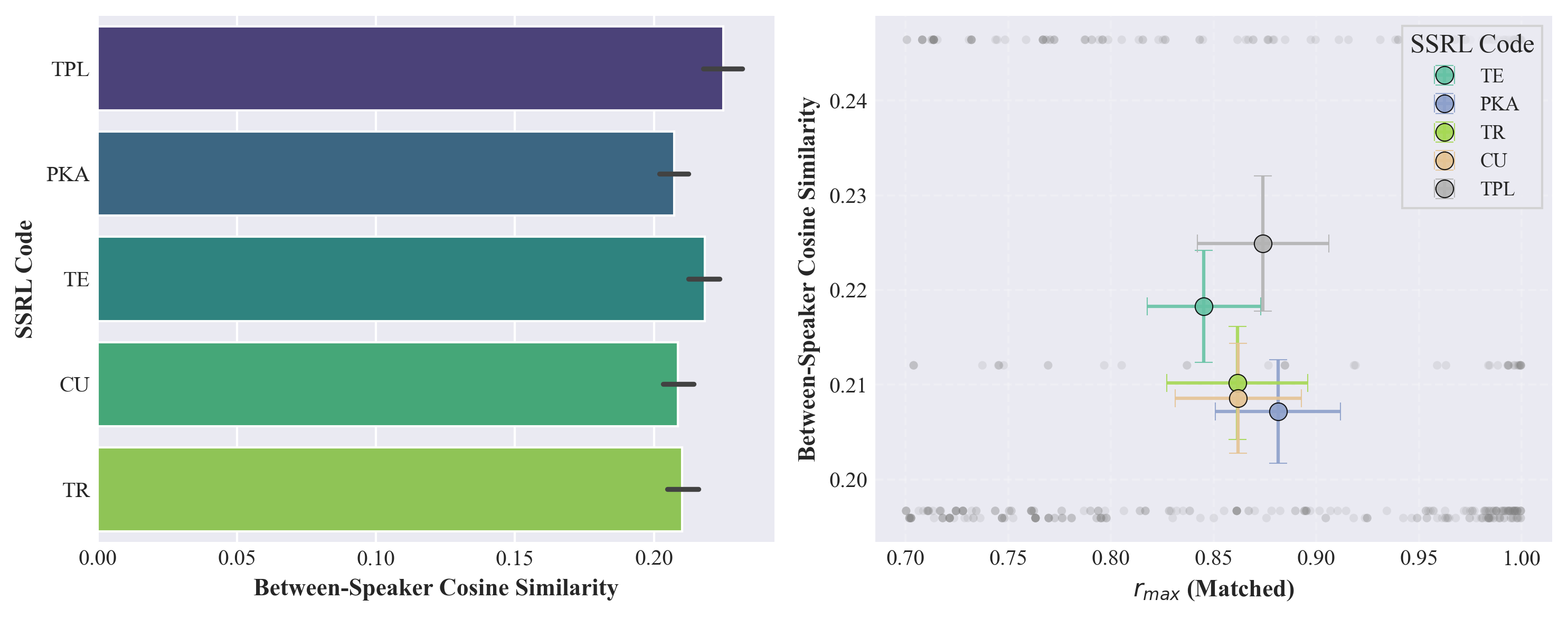}
\caption{Between-speaker similarity patterns. %
Task execution (TE)maintains higher similarity values, while prior knowledge activation (PKA) shows relatively lower alignment.}
\label{fig:between_similarity}
\end{figure}

Figures 1 \& 2 reveal consistent patterns across the two similarity measures. Task Execution (TE) consistently appears with higher similarity values in both within-speaker and between-speaker analyses, suggesting more predictable language patterns during executive actions towards the task. The scatter plots show a generally negative relationship between $r_{max}$ and cosine similarity, though with considerable variability across different interactions.%

\subsection{Qualitative Inspection on Pivotal Moments during PS Peaks}
Given that SSRL interactions did not differ significantly across PS levels, yet semantic variations were pronounced during both SSRL behaviors and PS peaks, we qualitatively inspected dialogues within high-PS segments. \setcounter{footnote}{0} Analysis revealed distinct patterns between groups that successfully diagnosed the patient and those that did not\footnote{\url{https://github.com/katherine-huang/AIED-2026/blob/main/quotes-physio-sync-peaks.png}}. In successful dyads (G1, G3, G4), PS peaks occurred during critical diagnostic phases and information clarification. For example, G1’s peaks involved intense synthesis (e.g., locating catecholamine data), while G3’s peaks coincided with evaluating results against expert findings (e.g., narrowing endocrine hypotheses). In G4, PS spiked when confirming data points aligned with their established mental model.

Conversely, in Group 2 (G2), which failed to reach the correct diagnosis, PS peaks were associated with repetitive looping over patient symptoms and expressions of uncertainty. Rather than progressing toward a solution, their high-coordination moments centered on unresolved hypotheses, such as questioning the relevance of the patient’s age or blood pressure management (“perhaps her blood pressure is not being appropriately managed... But of what?”). %

According to the participants' reflections, a consistent theme among successful diagnostic teams (G1, G4) was the shared enjoyment of confirmation and discovery. For these groups, PS peaks often aligned with moments of mutual curiosity and the validation of shared mental models. As noted by one participant: \begin{quote} ``We both said at one point, like, oh, this is actually fun... I think we both had the same view of the task. Like, we were both curious and wanting to learn from it and make it an interesting experience.'' (G4P1) \end{quote} Similarly, G1P2 reflected on the satisfaction of collaborative verification: It was fun to learn together, to confirm things together''. These statements suggest that for successful teams, physiological peaks mark moments of high-certainty alignment and positive affective resonance.

In contrast, the team that failed to reach a correct diagnosis (G2) also experienced PS peaks, but these were tied to the recognition of shared cognitive limits and uncertainty. One participant explicitly noted moments where they stuck: \begin{quote} ``I think at one point we both reached like a limit in our knowledge, but we both acknowledged that moment.'' (G2P2) \end{quote}

\section{Discussion and Implications}

\subsection{Identifying Maladaptive SSRL Behaviors}

Addressing RQ1, we integrated semantic embeddings of team dialogues with PS during collaborative medical diagnosis. The results suggest that PS peaks are more effective indicators of linguistic variation than sustained averages. These findings address a significant methodological gap in interpreting physiological synchrony within the context of medical practitioners' interaction~\cite{yan2025sync} and advance natural language processing application in AIED.

Furthermore, the quantitative relationship between SSRL behaviors, semantic similarity, and physiological measures sheds light on collaborative dynamics. The negative association found between physiological synchrony and linguistic similarity suggests that ``high-sync moments'' involve more exploratory, responsive, or varied language use rather than repetitive patterns. This finding supports that semantic alignment is related to physiological out-of-synchrony~\cite{lamsa2024learners}; specifically, highly synchronous moments appear to correspond with dynamic, less predictable linguistic exchanges. For example, while task execution follows more predictable discourse, cognitive and affective interactions (e.g., prior knowledge activation or social-emotional support) featured significantly higher linguistic variability. These findings indicate that physiological and linguistic coordination may operate independently rather than in parallel.

Overall, this analysis responds to the call for advanced multimodal learning analytics in AIED~\cite{di2025second} by combining novel language detection techniques with physiological signals and objective observations. The significance of this study is twofold: first, it highlights the semantic relationship with physiological measures, emphasizing the need for content-aware inspection in educational research. Second, the findings suggest that beyond verbal observations guided by the SSRL framework, investigating semantic dynamics using AI can facilitate customized team scaffolding. This is essential for identifying maladaptive regulation where targeted prompts and hints can be deployed to intervene and support the team.

\subsection{Toward a Multimodal Bio-Responsive Intelligent Tutoring}
Our results suggest that physiological synchrony peaks offer a promising signal for identifying when these pivotal transitions are occurring, while semantic features provide insight into what teams are doing during those moments. These findings support a future direction for collaborative tutoring systems: monitoring “team vital signs” through integrated physiological and verbal signals.

Qualitative results from RQ2 indicate that PS peaks are not universal markers of productivity; they reflect either shared discovery or stalled regulation, contingent on performance. Consequently, bio-responsive interventions should be contextualized using dialogue-level semantic and regulatory indicators. Future systems could (1) identify pivotal moments via synchrony peaks and (2) interpret these moments through SSRL-informed patterns. This dual approach enables adaptive support—such as prompting evidence-based evaluation or reflective coordination—only when teams struggle, avoiding interruptions during stable, productive execution. Beyond real-time scaffolding, these "team vital signs" facilitate post-task analytics. By generating synchrony maps that link physiological spikes to semantic themes and regulatory events, systems can help learners and educators visualize when teams aligned effectively and how regulation shaped diagnostic progress~\cite{sottilare2018designing,yan2025sync}.

\subsection{Summary and Conclusions}

This work advances AIED research toward systems that can recognize and support SSRL in real time. The study concludes that integrating physiological signals with semantic analysis provides a holistic view of ``team vital signs'' that is superior to analyzing either modality in isolation. Transient physiological synchrony peaks were the most reliable indicators of significant shifts in collaborative dialogue and SSRL. Notably, high synchrony was associated with lower semantic similarity, suggesting these moments involve more exploratory and varied language use. Qualitative findings clarified that these peaks represent ``pivotal moments,'' shared discovery for successful, but uncertainty for unsuccessful ones.

\bibliographystyle{splncs04}
\bibliography{references}

@article{sottilare2018designing,
  title={Designing adaptive instruction for teams: A meta-analysis},
  author={Sottilare, Robert A and Shawn Burke, C and Salas, Eduardo and Sinatra, Anne M and Johnston, Joan H and Gilbert, Stephen B},
  journal={International Journal of Artificial Intelligence in Education},
  volume={28},
  number={2},
  pages={225--264},
  year={2018},
  publisher={Springer}
}

@inproceedings{di2025second,
  title={The Second International Workshop on Multimodal Artificial Intelligence in Education (MAIEd’25)},
  author={Di Mitri, Daniele and Srivastava, Namrata and Fernandez Nieto, Gloria and Echeverria, Vanessa and Cobos, Ruth and Tudur Sadashiva, Ashwin and Spikol, Daniel and Wong, Kester Yew Chong and Zhou, Qi and Cukurova, Mutlu},
  booktitle={International Conference on Artificial Intelligence in Education},
  pages={292--299},
  year={2025},
  organization={Springer}
}

@article{lamsa2024learners,
  title={Learners’ Linguistic Alignment and Physiological Synchrony: Identifying Trigger Events that Invite Socially Shared Regulation of Learning},
  author={L{\"a}ms{\"a}, Joni and Edwards, Justin and Haataja, Eetu and Sobocinski, Marta and Pe{\~n}a, Paola and Nguyen, Andy and J{\"a}rvel{\"a}, Sanna},
  journal={Journal of Learning Analytics},
  volume={11},
  number={2},
  pages={197--214},
  year={2024}
}

@article{yan2025sync,
  title={In Sync or Out of Sync? Understanding Stress and Learning Performance in Collaborative Healthcare Simulations through Physiological Synchrony and Arousal},
  author={Yan, Lixiang and Ga{\v{s}}evi{\'c}, Dragan and Echeverria, Vanessa and Zhao, Linxuan and Jin, Yueqiao and Li, Xinyu and Martinez-Maldonado, Roberto},
  journal={International Journal of Artificial Intelligence in Education},
  pages={1--32},
  year={2025},
  publisher={Springer}
}

@article{martinez2021you,
  title={What do you mean by collaboration analytics? A conceptual model},
  author={Martinez-Maldonado, Roberto and Ga{\v{s}}evi{\'c}, Dragan and Echeverria, Vanessa and Fernandez Nieto, Gloria and Swiecki, Zachari and Buckingham Shum, Simon},
  journal={Journal of Learning Analytics},
  year={2021},
  publisher={Society for Learning Analytics Research}
}

@inproceedings{huang2025makes,
  title={What makes teamwork work? A multimodal case study on emotions and diagnostic expertise in an intelligent tutoring system},
  author={Huang, Xiaoshan and Wu, Haolun and Liu, Xue and Lajoie, Susanne P},
  booktitle={International Conference on Artificial Intelligence in Education},
  pages={44--52},
  year={2025},
  organization={Springer}
}

@inproceedings{huang2025linking,
  title={Linking Facial Recognition of Emotions and Socially Shared Regulation in Medical Simulation},
  author={Huang, Xiaoshan and Zhong, Tianlong and Wu, Haolun and Wang, Yeyu and Churchill, Ethan and Liu, Xue and Shaffer, David Williamson},
  booktitle={Companion Publication of the 2025 Conference on Computer-Supported Cooperative Work and Social Computing},
  pages={239--243},
  year={2025}
}

@article{huang2023exploring,
  title={Exploring the co-occurrence of students' learning behaviours and reasoning processes in an intelligent tutoring system: An epistemic network analysis},
  author={Huang, Xiaoshan and Li, Shan and Wang, Tingting and Pan, Zexuan and Lajoie, Susanne P},
  journal={Journal of computer assisted learning},
  volume={39},
  number={5},
  pages={1701--1713},
  year={2023},
  publisher={Wiley Online Library}
}

@article{jarvela2013new,
  title={New frontiers: Regulating learning in CSCL},
  author={J{\"a}rvel{\"a}, Sanna and Hadwin, Allyson F},
  journal={Educational psychologist},
  volume={48},
  number={1},
  pages={25--39},
  year={2013},
  publisher={Taylor \& Francis}
}

@article{malmberg2019we,
  title={Are we together or not? The temporal interplay of monitoring, physiological arousal and physiological synchrony during a collaborative exam},
  author={Malmberg, Jonna and Haataja, Eetu and Sepp{\"a}nen, Tapio and J{\"a}rvel{\"a}, Sanna},
  journal={International Journal of Computer-Supported Collaborative Learning},
  volume={14},
  number={4},
  pages={467--490},
  year={2019},
  publisher={Springer}
}

@article{huang2025examining,
  title={Examining socially shared regulation of learning in medical training: the interplay of heart rate change points on regulatory interactions},
  author={Huang, Xiaoshan and Nguyen, Andy and Lajoie, Susanne P},
  journal={European Journal of Psychology of Education},
  volume={40},
  number={3},
  pages={93},
  year={2025},
  publisher={Springer}
}

@article{liu2021collaborative,
  title={Collaborative learning quality classification through physiological synchrony recorded by wearable biosensors},
  author={Liu, Yang and Wang, Tingting and Wang, Kun and Zhang, Yu},
  journal={Frontiers in Psychology},
  volume={12},
  pages={674369},
  year={2021},
  publisher={Frontiers Media SA}
}

@inproceedings{borchers2025can,
  title={Can large language models match tutoring system adaptivity? a benchmarking study},
  author={Borchers, Conrad and Shou, Tianze},
  booktitle={International Conference on Artificial Intelligence in Education},
  pages={407--420},
  year={2025},
  organization={Springer}
}

@article{howley2016towards,
  title={Towards careful practices for automated linguistic analysis of group learning},
  author={Howley, Iris K and Ros{\'e}, Carolyn Penstein},
  journal={Journal of Learning Analytics},
  volume={3},
  number={3},
  pages={239--262},
  year={2016}
}

@article{koroteev2021bert,
  title={BERT: a review of applications in natural language processing and understanding},
  author={Koroteev, Mikhail V},
  journal={arXiv preprint arXiv:2103.11943},
  year={2021}
}

@article{zhang2026using,
  title={Using Large Language Models to Detect Socially Shared Regulation of Collaborative Learning},
  author={Zhang, Jiayi and Borchers, Conrad and Cohn, Clayton and Srivastava, Namrata and Snyder, Caitlin and Guo, Siyuan and Mohammed, Naveeduddin and Noh, Haley and Biswas, Gautam and others},
  journal={arXiv preprint arXiv:2601.04458},
  year={2026}
}

@article{borchers2025disentangling,
  title={Disentangling Learning from Judgment: Representation Learning for Open Response Analytics},
  author={Borchers, Conrad and Patel, Manit and Lee, Seiyon M and Botelho, Anthony F},
  journal={arXiv preprint arXiv:2512.23941},
  year={2025}
}

@inproceedings{carmon2023assessment,
  title={Assessment in Conversational Intelligent Tutoring Systems: Are Contextual Embeddings Really Better?},
  author={Carmon, Colin M and Hu, Xiangen and Graesser, Arthur C},
  booktitle={International Conference on Artificial Intelligence in Education},
  pages={121--129},
  year={2023},
  organization={Springer}
}

@inproceedings{chen2024multi,
  title={Multi-turn classroom dialogue dataset: Assessing student performance from one-on-one conversations},
  author={Chen, Jiahao and Liu, Zitao and Hou, Mingliang and Zhao, Xiangyu and Luo, Weiqi},
  booktitle={Proceedings of the 33rd ACM International Conference on Information and Knowledge Management},
  pages={5333--5337},
  year={2024}
}

@inproceedings{liu2024assessing,
  title={Assessing the potential and limits of large language models in qualitative coding},
  author={Liu, Xiner and Zhang, Jiayi and Barany, Amanda and Pankiewicz, Maciej and Baker, Ryan S},
  booktitle={International conference on quantitative ethnography},
  pages={89--103},
  year={2024},
  organization={Springer}
}

\end{document}